# Solving Poisson Equation by Genetic Algorithms

Khalid Jebari
LCS Laboratory, Faculty of Sciences, Mohammed V Agdal
Rabat Morocco

Mohammed Madiafi
Ben M'Sik faculty of Sciences
Mohammadia
Morocco

Abdelaziz El Moujahid
LCS Laboratory, Faculty of Sciences, Mohammed V Agdal
Rabat Morocco

## ABSTRACT
This paper deals with a method for solving Poisson Equation (PE) based on genetic algorithms and grammatical evolution. The method forms generations of solutions expressed in an analytical form. Several examples of PE are tested and in most cases the exact solution is recovered. But, when the solution cannot be expressed in an analytical form, our method produces a satisfactory solution with a good level of accuracy.

## General Terms
Artificial Intelligence, Algorithms.

## Keywords
Genetic algorithms, Evolutionary Computation, Poisson Equation, Grammatical Evolution

## 1. INTRODUCTION
Poisson equation is omnipresent in science, engineering and manufacturing. Several physical phenomena may be described by PE [1]. In electrostatics, the electric field E can be expressed in terms of an electric potential φ:

$$E = -\nabla \varphi \quad (1)$$

Where $\nabla$ is the divergence operator

The potential itself satisfies Poisson's equation:

$$\nabla^2 \varphi = -\frac{\rho}{\varepsilon_0} \quad (2)$$

Where $\nabla^2$ is Laplace operator, $\rho$ is the charge density and $\varepsilon_0$ is the permittivity of free-space. In Newtonian gravity, we can write the force $f$ exerted on a unit mass in terms of a gravitational potential ɸ:

$$f = -\nabla \varphi \quad (3)$$

The potential satisfies Poisson's equation:

$$\nabla^2 \varphi = 4\pi^2 G\rho \quad (4)$$

Where $\rho$ is the mass density, and G is the universal gravitational constant.

A series of problems in many scientific fields such as physics [2], chemistry [3], biology [4], economics [5], electrostatics [6] and semiconductor [7] can be modelled with the use of PE. The Poisson equations are also very important in computer vision. They arise in several computer vision fields, such as shape from shading, surface reconstruction, height from gradient and brightness based stereo vision [8].

Due to the importance of Poisson Equation, many methods have been proposed. In some cases, analytical solutions can be found or approximated by standard methods. However, in numerous cases these equations are nonlinear and are impossible to solve. In literature, many methods have been proposed for solving PE such as Runge Kutta methods [9], Predictor–Corrector [10], radial basis functions [11], artificial neural networks [12] and genetic programming [13].

In this paper, we have proposed a method based on Genetic Algorithms (GAs) and grammatical evolution. This method has also the advantage of not requiring the derivative of the objective function, which is a great advantage for problems whose objective function is not known in an analytic form.

GAs are stochastic methods that permit to find, in a reasonable amount of time, acceptable and satisfying solutions to challenging problems of search and optimization that are out of reach for conventional and deterministic methods. They are iterative heuristic procedures that imitate biological evolution as described by Darwin's theory of evolution [14]. Such heuristics have been proved effective in solving a variety of hard real-world problems in many application domains[ 15].

To design a genetic solution to any optimization problem, we first need to represent each candidate solution to the problem, called individual, by the mean of an abstract representation, called chromosome [15]. A function, called fitness, is necessary for assessing and comparing the relative quality of different solutions. Thus, starting from an initial population of randomly generated individuals. GA permits to evolve this population, throughout iterations called generations, toward better solutions according to rules of selection [16], crossover and mutation that simulate biological evolution.

Our method offers analytical form solutions. However, the variety of the basic functions involved is not a priori determined, rather is constructed dynamically as the solution procedure proceeds and evolve. The generation of solutions progress with using a grammatical evolution, governed by a grammar expressed in Backus Naur Form (BNF) [17].

The rest of the paper is organized as follows: in section 2 we give a brief presentation of grammatical evolution, in section 3 we describe in detail our method, in section 4 we present several experiments and in section 5 we present our conclusions and ideas for further work.

## 2. GRAMMATICAL EVOLUTION
Grammatical evolution (GE) is an evolutionary algorithm that can produce code in any programming language. GE has been applied successfully to problems such as symbolic regression [18], discovery of trigonometric identities [19], robot control [20], caching algorithms [21], and financial prediction [22].

The algorithm requires as inputs the BNF grammar definition of the target language and the appropriate fitness function. Chromosomes in grammatical evolution, in contrast to classical genetic algorithms, are expressed as vectors of integers. Each integer denotes a production rule from the BNF grammar. The algorithm starts from the start symbol of the grammar and gradually creates the string, by replacing non terminal symbols with the right hand of the selected production rule. The selection is performed in two steps:

- read an element from the chromosome (with value V)
- select the Rule according to the scheme

$$\text{Rule} = V \bmod R \qquad (1)$$

Where R is the number of rules for the specific non terminal symbol.

A BNF grammar is made up of the tuple N, T, P and S, where N is the set of all non-terminal symbols, T is the set of terminals, P is the set of production rules that map N to T, and S is the initial start symbol and a member of N. Where there are a number of production rules that can be applied to a non-terminal, a " | " (or) symbol separates the options. Using:

N = {<expr>, <op>, <operand>, <var>, <func>}

T = {1, 2, 3, 4, +, -, /, *, x, y, z}

S = {<expr>}

The process of replacing non terminal symbols with the right hand of production rules is continued until the end of chromosome has been reached. We can reject the entire chromosome or we can start over (wrapping event) from the first element of the chromosome if threshold of the number of wrapping events is reached, the chromosome is rejected by assigning to it a large fitness value. In our approach, threshold of the number of wrapping events is equal to 2.

As well we programmed in C++ language, we used a small part of the C programming language grammar as we can see an example in Table II, we also construct 4 Radial Basis Functions (RBF), see Table I, that we added to the C++ standard functions. We note that the numbers in parentheses denote the sequence number of the corresponding production. Below is a BNF grammar used in our method:

S ::= <expr>           (0)

<var> ::= x            (0)
      | y              (1)
      | z              (2)

<operand> ::= 0        (0)
           | 1         (1)
           | 2         (2)
           | 3         (3)
           | 4         (4)
           | 5         (5)
           | 6         (6)
           | 7         (7)
           | 8         (8)
           | 9         (9)
           | <var>     (10)

<op> ::= +             (0)
      | -              (1)
      | *              (2)
      | /              (3)

<func> ::= sin         (0)
        | cos          (1)
        | exp          (2)
        | log          (3)
        | sqrt         (4)
        | BRF1         (5)
        | BRF2         (6)
        | BRF3         (7)
        | BRF4         (8)

<expr> ::= <expr> <op> <expr>    (0)
        | ( <expr> )             (1)
        | <func> ( <expr> )      (2)
        | <operand>              (3)

**Table 1. Radial Basis Functions**

| Name Used | Function Name | Definition |
|-----------|---------------|------------|
| BRF1 | Gaussian | $\exp(-cr^2)$ |
| BRF2 | Hardy Multiquadratic | $\sqrt{c^2 + r^2}$ |
| BRF3 | Inverse Multiquadratic | $\sqrt{\dfrac{1}{c^2 + r^2}}$ |
| BRF4 | Inverse Quadratic | $\dfrac{1}{c^2 + r^2}$ |

Consider the chromosome C, the steps of the mapping procedure are listed in Table 2. The result of these steps is the expression sqrt(3*x)

**Table 2. An example for executing procedure**

| String_BNF | Chromosome | Operation |
|---|---|---|
| <expr> | 10,4,8,15,3,6,19,21,9 | 10 mod 4=2 |
| <func>(<expr>) | 4,8,15,3,6,19,21,9 | 4 mod 9=4 |
| sqrt(<expr>) | 8,15,3,6,19,21,9 | 8 mod 4=0 |
| sqrt(<expr><op><expr>) | 15,3,6,19,21,9 | 15 mod 4=3 |
| sqrt(3*<expr>) | 19,21,9 | 19 mod 4=3 |
| sqrt(3*<operand>) | 21,9 | 21 mod 11=10 |
| sqrt(3*<var>) | 9 | 9 mod 3=0 |
| sqrt(3*x) | | |

## 3. METHOD DESCRIPTION

In order to conceive a genetic solution to PE we have to determine the encoding method. Then the fitness function is used for assessing and comparing the solutions candidates for PE. Thus, starting from an initial population of randomly generated individuals, we evolved this population toward better solutions according to the rules of selection strategy, crossover and mutation. The details are as follows

**Encoding Method:** the ordinal encoding scheme was used in the proposed method. Under this scheme, a serial number is assigned to each gene from 0 to s where s=50.

**Initial population size**: generally, the initial population size can be determined according to the complexity of the solved problem. A larger population size will reduce the search speed of the GA, but it will increase the probability of finding a high quality solution. The initialization of the population of chromosomes is generated by a random process. Each chromosome is represented as permutation $j_1, j_2, ... , j_s$ of 1, 2, ... , s. The process does not yield any illegal chromosome, i.e. the alleles of each gene of the chromosome are different from each others.

**Fitness function:** the fitness function is a performance index that it is applied to judge the quality of the generated solution of PE.

We only consider the PE in two variables with Dirichlet boundary conditions.

The generalization of the process to other types of boundary conditions and higher dimensions is possible with our method. The PE is expressed in the form:

$$\nabla^2 u(x,y) - \rho(x,y) = 0 \qquad (5.a)$$

Or

$$f\left(x,y,u(x,y),\frac{\partial^2}{x^2}u(x,y),\frac{\partial^2}{y^2}u(x,y)\right) = 0 \qquad (5.b)$$

With $x \in [a,b]$ $y \in [c,d]$

Suppose that u(x, y) satisfies mixed boundary conditions in the x-direction: *i.e.*,

u(0,y)-$\gamma_a$(y)=0

u(1,y)-$\gamma_b$(y)=0;

Furthermore, suppose that u(x, y) satisfies the following simple Dirichlet boundary conditions in the y-direction:

u(x,0)-$\beta_c$(x)=0;

u(x,1)-$\beta_d$(x)=0;

The steps for evaluating chromosome $C_i$ by fitness function are [13]:

1. Choose $T_x$ equidistant points in [a,b]
   $T_y$ equidistant points in [c,d],

   we also choose $T_x=T_y=T$;

2. Construct a solution $S_i$ from $C_i$ by using grammatical evolution described earlier;

3. Calculate
$$E_r(S_i) = \sum_{j=1}^{T^2} f\left(x_j, y_j, S_i(x,y), \frac{\partial^2}{x^2}S_i(x,y), \frac{\partial^2}{y^2}S_i(x,y)\right)$$

4. Calculate the penalty value satisfying each Dirichlet boundary condition:

$$P_a(S_i) = \sum_{j=1}^{T} S_i(a,y_i) - \gamma_a(y_i) \quad (7)$$

$$P_b(S_i) = \sum_{j=1}^{T} S_i(b,y_i) - \gamma_b(y_i) \quad (8)$$

$$P_c(S_i) = \sum_{j=1}^{T} S_i(x_i,c) - \beta_c(x_i) \quad (9)$$

$$P_d(S_i) = \sum_{j=1}^{T} S_i(x_i,d) - \beta_d(x_i) \quad (10)$$

5. the fitness is :
$$f(c_i) = E_r + (P_a + P_b + P_c + P_d)(S_i) \quad (11)$$

**Selection operator:** in the selection operation, the chromosome with the lower fitness function value will have a higher probability to reproduce the next generation.

The aim of this operation is to choose a good chromosome to achieve the goal of gene evolution. The most commonly used method is Tournament Selection. In this study, we used a modified Tournament Selection, which guards in each iteration the best individual. The following pseudo code summarizes the Selection Method.

```
// N: population size

T_alea: array of integer containing the indices of individuals in
the population

T_ind_Winner : an array of individuals indices 's who will be
selected

L_sorted : a list of all individual indices sorted in decreasing
fitness values

l = 0

k = 0

For (i=0; i<k; i++)
{
        Shuffle T_alea ;

        For (j=0; j<N; j=j+k+1)
        {
                C_1 = T_alea(j);

                For (m=1; m<k; m++)
                {
                        C_2 = T_alea(j+m);

                        if f(C_1)< f(C_2)  C_1 = C_2
                }
        // f(C_i): Fitness of individual C_i
        }

        T_ind_Winner(l) = C_1

        T_ind_Winner(l+1) = L_sorted (k)

        l=l+2;

        k=k+1
        }
}
```

**Fig 1. Tournament Selection Modified**

**Crossover operation**: This operation aims to combine two parent chromosomes to generate better child chromosomes, by crossover rate decision.

In our study, a novel Homologous crossover [23] operator is used, which is inspired by molecular biology and proceeds as follows:

1. For each individual of the population, a history of the rules in the grammar BNF is stored.
2. The histories of the two individuals for crossover are aligned.
3. The history for each parent is treated from the left with sequential manner. The region of similarity is marked if the rules selected for the both parents are identical.
4. At the limits of the region of similarity, the first two crossover points are picked. These points are the same on both parents
5. The two second crossover points are then selected from the regions of dissimilarity by respecting the following steps:
   5. 1 from the dissimilarity region, the crossover point is selected randomly in the first parent;
   5. 2 on the second parent, the crossover point picked respects the rule that a gene is associated with the same type of non-terminal as the gene related the crossover point on the first parent;
   5. 3 this gene is established by generating randomly a crossover point and then the process searches incrementally, nearby this point until an appropriate point is found;
   5. 4 once the second crossover point is found, the process applies the simple two points crossover;
   5. 5 if no point is found in the second parent, the crossover process is unsuccessful and a new initial crossover point is randomly selected in the first parent;

In this research, the probability of crossover is 0.7

**Mutation operation:** we used Inversion Mutation, the inversion mutation [24] is randomly select a sub string_BNF, removes it from the string_BNF and inserts it in a randomly selected position. However, the sub string_BNF is inserted in reversed order. The mutation operation will create some new individuals that might not be produced by the reproduction and crossover operations.

Generally, a lower probability of mutation can guarantee the convergence of the GA, but it may lead to a poor solution quality. On the other hand, a higher probability of mutation may lead to the phenomenon of a random walk for the GA. In this research, the probability of mutation is set to be 0.1.

**Stop criterion**: the genetic algorithm repeatedly runs the reproduction, crossover, mutation, and replacement operations until it meets the stop criterion. The stop criterion is set to be 1000 generations, because this criterion can obtain satisfied solution

## 4. EXPERIMENTAL RESULTS
The proposed method was tested on a series of PEs, with two and three variables and Dirichlet boundary conditions. These test functions are listed subsequently and they have been used in the experiments performed in [1,2,13]. In the following, the PEs are presented with the exact solutions:

1.
$$\nabla^2 u_1(x,y) = -32\pi^2 \sin(4\pi\pi)\sin(4\pi\pi)$$

$x, y \in [-1,1]$
$u_1(\pm 1, y) = 0$
$u_1(x, \pm 1) = 0$

The exact solution is

$$u_1(x,y)=\sin(4\pi x)\sin(4\pi y)$$

2.
$$\nabla^2 u_2(x,y) = 2(x^2 + y^2) + 2x(y^2-1) + (x^2-1)(y^2+1)\exp(x+y)$$

$x, y \in [-1,1]$
$u_2(\pm 1, y) = 0$
$u_2(x, \pm 1) = 0$

The exact solution is

$$u_2(x,y)=(x^2-1)(y^2-1)\exp(x+y)$$

3.
$$\nabla^2 u_3(x,y) = 6xy(1-y) - 2x^3$$
$x, y \in [0,1]$

$u_3(x,0)=0; u_3(x,1)=0$

$u_3(0,y)=0; u_3(1,y)=y(y-1)$

The exact solution is

$$u_3(x,y)=y(y-1)x^3$$

4.
$$\nabla^2 u_4(x,y,z) = 6$$
$x, y, z \in [0,1]$

$u_4(0,y,z)=y^2+z^2$; $u_4(1,y,z)=1+y^2+z^2$

$u_4(x,0,z)=x^2+z^2$; $u_4(x,1,z)=1+x^2+z^2$

$u_4(x,y,0)=x^2+y^2$ ; $u_4(x,y,1)=1+x^2+z^2$

The exact solution is

$$u_4(x,y,z)= 1+x^2+y^2+z^2$$

We used for the population size a number in [200,1000], the chromosome length is 50. The number of equidistant points $T_x=T_y=T=100$.

The experiments were performed on a core duo 2400 Mhz, running in Debian Linux. Table III summarizes the results:

**Table 3. Solution founded by the proposed algorithm**

| Problem | Optimum Fitness Value | Solution Founded |
|---|---|---|
| u1(x,y) | 2.8 x 10$^{-7}$ | sin(4πx)(sin(4πy)) |
| u2(x,y) | 1.2 x 10$^{-8}$ | (y$^2$-1)exp(x+y)(x$^2$-1) |
| u3(x,y) | 3.5 x 10$^{-8}$ | y(y-1)x$^3$ |
| u4(x,y,z) | 4.1 x 10$^{-6}$ | 1+x$^2$+y$^2$+z$^2$ |

We consider now a problem of sinh Poisson Equation (shPE), governed by the relation:

$$\nabla^2 u_5(x,y) + \sinh(u(x,y)) = 0$$

The shPE uses in the studies of the most probable state in viscid two-dimensional flows in fluids and plasmas. Exact solution in analytical form is found by Mallier–Maslowe [25]:

$$u_5(x,y) = 4 * \tanh^{-1}\left(\frac{\cos(\sqrt{2}x)}{\sqrt{2}\cosh(y)}\right)$$

Our method found the following analytic solution:

$$u_{m5}(x,y) = 4 * \left[\frac{1}{1+\exp\left(\frac{\cos(\sqrt{2}x)}{\exp(y)+\exp(-y)}\right)}\right]$$

$$+ 4 * \left[\frac{1}{-1+\sqrt{2}*\exp\left(\frac{\cos(\sqrt{2}x)}{\exp(y)+\exp(-y)}\right)}\right]$$

The difference between the trial solution $u_m(x; y)$ and the exact solution found by Mallier–Maslowe is shown

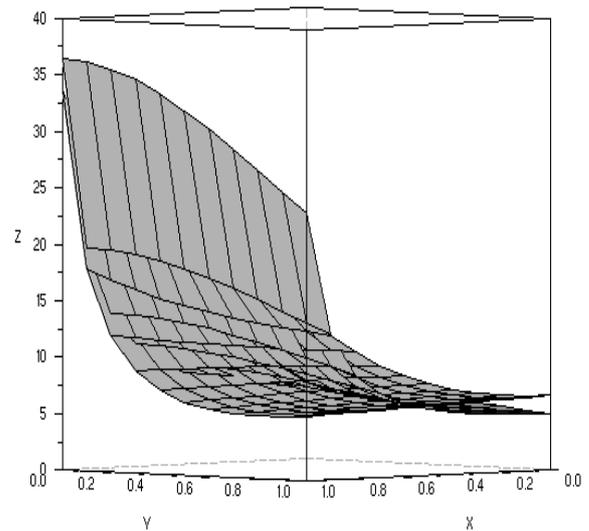

**Fig 2: Difference between Analytical Solution $u_5(x,y)$ and $u_{m5}(x,y)$**

## 5. CONCLUSION

Solving PE is a major concern, both with respect to the complexity of the task and to its relevance in the mathematical representation of problems in almost all the branches of sciences. Taking into account the algorithmic character of many of the existing solving methods, it is clear that GAs can play an important role in the progress of solving PE.

In this context, the goal of the material presented was to contribute both to the discussion of the GAs role in this kind of problems and to the implementation of this algorithm using the grammatical evolution.

In this paper, we have developed an algorithm, based on a real coded genetic algorithm and grammatical evolution for solving PE.

The GA based method was found to produce trial solutions and minimize an associated error. If the proposed method cannot produce an exact solution, it will be able to recover an approximant form.

Although we have only considered solving PE, the method can be applied to more general linear and nonlinear partial differential equations.

Concerning the possible extensions of this work, they can relate to the interface, the computational capabilities and the mathematical methods. Some of the possible improvements falling into these categories are:

- the grammar used can be further developed and enhanced.
- to give the User the option of building a family of Partial differential equations through a friendly interface;
- to complete the implementation of a reasonable collection of standard methods;
- to introduce techniques for working with systems of coupled linear PDEs;
- to introduce symmetry methods for solving PDEs.

These improvements are expected to be included in future works.